%% file: main.tex
\documentclass[runningheads]{llncs}
\usepackage{todonotes}
\usepackage{wrapfig}

\usepackage{etoolbox}
\BeforeBeginEnvironment{wrapfigure}{\setlength{\intextsep}{0pt}}

\usepackage[acronym,nonumberlist]{glossaries}

\usepackage{amsmath}
\usepackage{amssymb}
\usepackage{xspace}
\usepackage{paralist}
\input{acronyms}

\usepackage{graphicx}
\usepackage{tikz}
\usetikzlibrary{arrows,automata}
\usepackage{pgfplots}
\pgfplotsset{compat=newest}
\usepgfplotslibrary{groupplots}
\usepgfplotslibrary{dateplot}
\usepackage{multirow}
\usepackage{float}
\usepackage{svg}
\usepackage{wrapfig}
\usepackage{pifont}
\usepackage{framed}

\usepackage{amsfonts}
\usepackage{amsmath}
\usepackage{amssymb}

\usepackage[noend]{algpseudocode}
\usepackage{algorithm}
\makeatletter
\renewcommand{\ALG@beginalgorithmic}{\scriptsize}
\makeatother
\algrenewcommand\alglinenumber[1]{\scriptsize #1:}

\input{vertical_alg_lines}

\usepackage{etoolbox}
\usepackage{url}
\usepackage{stmaryrd}
\usepackage{pgfplotstable}
\pgfplotsset{compat=1.11,
    /pgfplots/ybar legend/.style={
    /pgfplots/legend image code/.code={%
       \draw[##1,/tikz/.cd,yshift=-0.25em]
        (0cm,0cm) rectangle (3pt,0.8em);},
   },
}

\input{defs}

\usepackage{hyperref}

\begin{document}
\title{Reinforcement Learning\\ under Partial Observability\\ Guided by Learned Environment Models}
%
%
\author{
Edi Mu\v{s}kardin\inst{1,2}\orcidID{0000-0001-8089-5024} \and 
Martin Tappler\inst{2,1}\orcidID{0000-0002-4193-5609} \and
Bernhard~K.~Aichernig\inst{2}\orcidID{0000-0002-3484-5584} \and
Ingo Pill\inst{1}\orcidID{0000-0002-8420-6377} 
}

\authorrunning{Mu\v{s}kardin et al.}
\titlerunning{Reinforcement Learning in Partially Observable Environments}
%
\institute{
Silicon Austria Labs, TU Graz - SAL DES Lab, Graz, Austria
\and
Graz University of Technology, Institute of Software Technology, Graz, Austria
}
\maketitle              
\begin{abstract}

In practical applications, we can rarely assume full observability of a system's environment, despite such knowledge being important for determining a reactive control system's precise interaction with its environment. 
Therefore, we propose an approach for reinforcement learning (RL) in partially 
observable environments. While assuming that the environment behaves
like a partially observable Markov decision process with known discrete
actions, we assume 
no knowledge about its structure or transition probabilities.

Our approach combines Q-learning with IoAlergia, a method for learning Markov decision processes (MDP).
By learning MDP models 
of the environment from
episodes of the RL agent,
we enable RL in partially observable domains without
explicit, additional memory to track
previous interactions for dealing with ambiguities stemming from partial observability. 
We instead provide RL with additional observations
in the form of abstract environment states by simulating new
experiences on learned environment models to track the 
explored states. 
In our evaluation we report on the validity of our approach and its promising performance in comparison to six state-of-the-art deep RL techniques with recurrent neural networks and fixed memory.



\end{abstract}
\setlength{\intextsep}{0mm}

\section{Introduction}
\label{sec:intro}
\input{sections/intro}
\section{Preliminaries}
\label{sec:prelim}
\input{sections/prelim}

\section{\poqlearning: RL Assisted by Automata Learning}
\label{sec:method}
\input{sections/method}
\section{Evaluation}
\label{sec:eval}
\input{sections/eval}

\section{Related Work}
\label{sec:rel_work}
\input{sections/relatedWork}

\section{Conclusion}
\label{sec:conclusion}
\input{sections/conclusion}
\subsubsection{Acknowledgments.}
This work has been supported by the "University SAL Labs" initiative of Silicon Austria Labs (SAL) and its Austrian partner universities for applied fundamental research for electronic based systems.

\apptocmd{\sloppy}{\hbadness 10000\relax}{}{}
\bibliographystyle{splncs04}
\bibliography{references.bib}


\end{document}

%% file: acronyms.tex
\newacronym[firstplural=Markov decision processes (MDPs),longplural=Markov decision processes]{MDP}{MDP}{Markov decision process}
\newacronym[plural=SMMs,firstplural=stochastic Mealy machines (SMMs)]{SMM}{SMM}{stochastic Mealy machine}
\newacronym{DTMC}{DTMC}{discrete time Markov chain}

\newacronym[plural=DFAs,firstplural=deterministic finite automata (DFAs)]{DFA}{DFA}{deterministic finite automaton}
\newacronym[plural=NFAs,firstplural=non-deterministic finite automata (NFAs)]{NFA}{NFA}{nondeterministic finite automaton}
\newacronym[plural=SULs,firstplural=systems under learning (SULs)]{SUL}{SUL}{system under learning}
\newacronym{MAT}{MAT}{minimally adequate teacher}
\newacronym{FSM}{FSM}{finite-state machine}
\newacronym{ONFSM}{ONFSM}{observable non-det\-er\-minis\-tic finite-state machine}
\newacronym{IOPT}{IOPT}{input-output prefix tree}
\newacronym[firstplural=partially observable Markov decision processes (POMDPs),longplural= partially observable Markov decision processes]{POMDP}{POMDP}{partially observable Markov decision process}
\newacronym[plural=BMDPs,firstplural=Belief MDPs (BMDPs)]{BMDP}{BMDP}{Belief MDP}
\newacronym[plural=DLMDPs,firstplural=Deterministic Labeled MDPs (DLMDPs)]{DLMDP}{DLMDP}{Deterministic Labeled MDP}

%% file: defs.tex
\newcommand{\poqlearning}{Q\textsuperscript{A}-learning}\xspace

\newcommand{\Dist}{\mathit{Dist}}
\newcommand{\supp}[1]{\textit{supp}(#1)}
\newcommand{\epsal}{\epsilon_\textsc{al}}

\newcommand{\Ret}{\ensuremath{\mathit{Ret}}}

\newcommand{\sample}{\mathcal{T}}
\newcommand{\rewsample}{\mathcal{RT}}

\newcommand{\IOALERGIA}{\textsc{IoAlergia}}

\algrenewcommand\Return{\State \algorithmicreturn{} }%

\makeatletter
\providecommand\theHALG@line{\thealgorithm.\arabic{ALG@line}}
\newcounter{algorithmicH}
\let\oldalgorithmic\algorithmic
\renewcommand{\algorithmic}{%
  \stepcounter{algorithmicH}
  \oldalgorithmic}
\renewcommand{\theHALG@line}{ALG@line.\thealgorithmicH.\arabic{ALG@line}}
\makeatother



%% file: sections/intro.tex
Reinforcement learning (RL) enables the automatic creation of controllers in stochastic environments
through exploration guided by rewards. Partial observability presents a challenge to RL, 
which naturally arises in various control problems. Unreliable or inaccurate sensor readings
may provide incomplete state information, e.g., static images provided by visual sensors 
do not capture the agent's movement trajectory and speed. Formally, partial observability 
occurs when observations of the environment do not allow to deduce the environment state directly. 
In such settings, optimal control based on observations only is generally not possible.  

For this reason, RL methods often include some form of memory to cope with partial 
observability, such as the hidden state in recurrent neural networks~\cite{DBLP:conf/aaaifs/HausknechtS15} or fixed-size memory obtained by concatenating 
previous observations~\cite{DBLP:journals/corr/MnihKSGAWR13}.
In this paper, we propose a method for RL in partially observable environments
that combines Q-learning~\cite{Watkins1992} with \IOALERGIA~\cite{DBLP:journals/ml/MaoCJNLN16},
a technique for learning deterministic labeled \glspl*{MDP}. 
With \IOALERGIA, we regularly learn and update MDPs based on the experiences of the RL agent.
The learned MDPs approximate the dynamics of the \gls*{POMDP} underlying the environment 
and their states extend the observation space of Q-learning. To enable this extension, 
we trace every step of the RL agent on the most recently learned MDP and add the explored MDP state
as an observation. Hence, we provide memory by tracking learned environmental states.
With this approach, we follow the tradition of state estimation in RL
under partial observability~\cite{DBLP:conf/aaai/Chrisman92,DBLP:conf/icml/McCallum93}. In comparison to earlier work, we overcome strict assumptions on the underlying~\gls*{POMDP}, such as knowledge about the number of states, e.g., criticized by Singh et al.~\cite{DBLP:conf/icml/SinghJJ94}. 

Our contributions comprise: (1) an approach for RL under partial observability aided 
by automata learning, which we term \poqlearning, (2) an implementation of the approach in 
an environment conforming to the OpenAI gym interface~\cite{DBLP:journals/corr/BrockmanCPSSTZ16}, and 
(3) its evaluation against three baseline deep RL methods with fixed memory 
and three RL methods with LSTMs providing memory.


{\it Structure.} In Sect.~\ref{sec:prelim}, we introduce preliminaries like passive learning of stochastic automata. We present our method for reinforcement learning in partially observable environments in Sect.~\ref{sec:method}, followed by a corresponding evaluation in Sect.~\ref{sec:eval}. After discussing related work in Sect.~\ref{sec:rel_work}, we conclude by summarizing our findings and providing an outlook on future work in Sect.~\ref{sec:conclusion}.


%% file: sections/prelim.tex

\subsection{Models}
In RL, we commonly assume that the environment behaves like an \gls{MDP} (see Def.~\ref{def:MDP}). An agent observes the environment's state and based on that reacts by choosing from a given set of actions---causing a probabilistic state transition. 
\begin{definition}[\acrfullpl{MDP}]\label{def:MDP}
 A \acrfull{MDP} is a tuple $\mathcal{M} = (S,s_0,A,\delta)$, where $S$ is a finite set of states, $s_0 \in S$ is the initial state,
 $A$ is a finite set of actions, $\delta : S \times A \rightarrow \Dist(S)$ is a probabilistic transition function. 
\end{definition}
Please note that for a simplified presentation, we assume \glspl*{MDP} to support all actions in all states, s.t. $\delta$ is total. In our work, we consider settings where the agent cannot observe the environment directly, but where it has only limited information---like a room's number, but not its position in the room (see Fig.~\ref{fig:officeWorld})
---and where we assume discrete states as well as finite action and state spaces. Such scenarios are commonly modeled as \acrfullpl{POMDP} (Def.~\ref{def:POMDP}), see, e.g., Bork et al.~\cite{DBLP:conf/atva/BorkJKQ20}. Alternative \gls*{POMDP} definitions including probabilistic observation functions can also be 
handled~\cite{DBLP:conf/icra/ChatterjeeCGK15}.

\begin{definition}[Partially observable Markov decision processes]\label{def:POMDP}
 A \acrfull{POMDP} is a triple $(\mathcal{M}, Z, O)$, where $\mathcal{M} = (S,s_0,A,\delta)$ is the underlying \gls*{MDP}, $Z$
 is a finite set of observations and $O : S \rightarrow Z$ is the observation function. 
\end{definition}

\begin{example}[Hot Beverage POMDP]
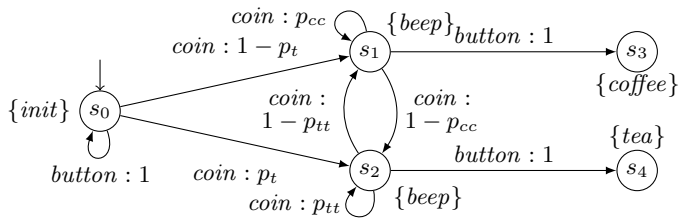
\begin{figure}[t]
    \centering
\input{figures/coffee_pomdp}
    \caption{A POMDP producing hot beverages.}
    \label{fig:coffee_pomdp}
\end{figure}
Fig.~\ref{fig:coffee_pomdp} shows a POMDP of a vending machine 
that, depending on parameterized probabilities, produces either tea or coffee. 
For the individual states $s_i$, we show the respective observation in curly braces. For each probabilistic transition reported (for brevity we ommit the transitions from $s_3$ and $s_4$, but they would loop back to $s_0$ for any action) we show a corresponding edge, labeled by the action and the transition's probability. While the parameterized probabilities will become more important later on, let us for now assume $p_t = 0.5$, $p_{cc} = 0.9$, and $p_{tt} = 0.1$. In the initial state $s_0$, for the action $\mathit{coin}$, we would now progress to \emph{either} $s_1$ \emph{or} $s_2$, but where the resulting observation would be $\mathit{beep}$ for both. Pressing a $\mathit{button}$, we would then move to $s_3$ or $s_4$ receiving either a $\mathit{coffee}$ or $\mathit{tea}$. Alternatively, we can add another $\mathit{coin}$ to move to $s_1$ with a probability of $0.9$ for increasing the chances to get a $\mathit{coffee}$.
\end{example}


\noindent
\emph{Paths, Traces \& Policies.}
The interaction of an agent with its environment can be described by a \emph{path} that is defined by an alternating sequence of states and actions $s_0 \cdot a_1 \cdot s_1, \cdots, s_n$ starting in the initial state. We denote the set of paths in an \gls*{MDP} $\mathcal{M}$ by $Paths_\mathcal{M}$. For partially observable scenarios, \emph{traces} basically replace the states in a path with the corresponding observations. We correspondingly lift observation functions $O$ to paths, applying $O$ on every state to derive trace $O(p) = L(s_0) \cdot a_1 \cdot L(s_1)$ from path $p = s_0 \cdot a_1 \cdot s_1$.
An agent selects actions based on a \emph{policy} that is a mapping from $Paths_\mathcal{M}$ to distributions over actions $\Dist(A)$. If a policy $\sigma$ depends only on the current state, we say that $\sigma$ is memoryless. 
With policies relating to action choices, an \gls*{MDP} controlled by a policy defines a probability distribution over paths. 


\emph{Rewards} define the crucial feedback an agent needs during learning for judging whether the actions it chose were ``good or bad''. That is, the goal is to learn a policy that maximizes the reward. To this end, we consider a reward function $R : S \rightarrow \mathbb{R}$ that returns a real value\footnote{Alternative definitions including actions are possible as well.}. For a path $p = s_0 \cdot a_1 \cdot s_1, \cdots, s_n$, we can define a discounted cumulative reward at time step $t$ as $\Ret(p,t) = \sum_{i=0}^{n-t-1} \gamma^i R(s_{t+i+1})$, taking a (time) discount factor $\gamma$ into account. For a memoryless policy $\sigma$, we can define a value function for a state $s$ as $v_\sigma(s) = \mathbb{E}_\sigma\left[\Ret(p,t) \mid s_t = s\right]$. To accommodate partial observability, we define reward-observation traces that extend traces with rewards, e.g., $rt = RO(p) = O(s_0) \cdot R(s_0) \cdot a_1 \cdot O(s_1) \cdot R(s_1), \cdots, O(s_n) \cdot R(s_n)$ for path $p=s_0 \cdot a_1 \cdot s_1, \cdots, s_n$. 


Note that in RL, we usually consider memoryless policies---enabled by the assumption of a Markovian environment (modeled as MDP) which guarantees that there is an optimal, memoryless policy for maximizing the reward. With partial observability it is impossible to precisely identify the current state (consider the Hot Beverage example and $s1$ vs. $s2$), meaning that creating optimal policies for \glspl*{POMDP} entails taking the history of previous actions into account---rendering the problem non-Markovian.
Alternatively, deriving policies under partial observability can be approached by creating
belief-\glspl{MDP} from \glspl*{POMDP}~\cite{DBLP:conf/aaai/CassandraKL94}. 


\emph{Belief-MDPs \& Deterministic Labeled MDPs.}
\glspl{DLMDP} feature an observation function and adhere to a specific determinism property that guarantees that any possible (observation) trace reaches exactly one state. \glspl{BMDP} are special \glspl*{DLMDP} that represent the dynamics of a \gls*{POMDP} and are defined over so-called belief states. These belief states (beliefs for short) describe probability distributions over states in a \gls*{POMDP}, i.e., over those states that the paths relating to a trace would reach in the \gls*{POMDP}. That is, for any given trace, a \gls*{BMDP} progresses to a unique state that in turn defines a distribution over possible \gls*{POMDP} states.

\begin{definition}[Deterministic Labeled MDPs]
 A deterministic labeled MDP is a triple $(\mathcal{M}, Z, O)$, 
 where $\mathcal{M} = (S,s_0,A,\delta)$ is the underlying \gls*{MDP}, 
 $Z$ is a set of  observations, and $O$ is an observation function,
 satisfying 
 \begin{align*}
  \forall s,s',s'' \in S, \forall a \in A\colon &\delta(s,a,s') > 0 \land \delta(s,a,s'') > 0 \land O(s') = O(s'') \\
  & \implies s' = s''.
 \end{align*}
\end{definition}
%
We introduce \glspl*{BMDP} with some auxiliary definitions: 
Let $P = (\mathcal{M}, Z, O)$ be a \gls*{POMDP} over an MDP $\mathcal{M} = (S,s_0,A,\delta)$. This defines the beliefs as the set $B= \{ \mathbf{b} \in \Dist(S) | \forall s,s' \in \supp{\mathbf{b}}: O(s) = O(s')\}$, where $\supp$ returns the support of a probability distribution. Then the probability of observing $z \in Z$ after executing $a\in A$
in $s\in S$ is defined as $\mathbf{P}(s,a,z) = \sum_{s' \in S, O(s') = z} \delta(s,a,s')$ and in a belief (state) $\mathbf{b}$ it is $\mathbf{P}(\mathbf{b},a,z) = \sum_{s \in S} \mathbf{b}(s) \cdot \mathbf{P}(s,a,z)$. If $O(s') = z$, the subsequent belief update is defined as $\llbracket\mathbf{b} | a, z \rrbracket(s') = \frac{\sum_{s \in S} \mathbf{b}(s) \cdot \delta(s,a,s')}{\mathbf{P}(\mathbf{b},a,z)}$. 

\begin{definition}[Belief MDPs]
The \gls*{BMDP} for a \gls*{POMDP} $P$ as of Def.~\ref{def:POMDP} is a \gls{DLMDP} $(\mathcal{M}_B, Z, O_B)$ over an MDP
$\mathcal{M}_B = (B,{s_0}_B,A,\delta_B)$, with $B$ as defined above, 
${s_0}_B = \{s_0 \mapsto 1\}$,  $O_B(\mathbf{b}) = O(s)$ for an $s \in \supp{\mathbf{b}}$, and 
\begin{equation*}
 \delta_B(\mathbf{b}, a, \mathbf{b'}) = \begin{cases}
                                         \mathbf{P}(\mathbf{b},a,O(\mathbf{b'})) & \text{ if } \mathbf{b'} = \llbracket\mathbf{b} | a, O(\mathbf{b'}) \rrbracket, \\
                                         0 &\text{ otherwise.}
                                        \end{cases}
\end{equation*}
\end{definition}
\glspl*{BMDP} allow to synthesize policies under partial observability, i.e., it was shown that an optimal policy
for a \gls*{BMDP} is optimal also for the corresponding\\ \gls*{POMDP}~\cite{DBLP:conf/aaai/CassandraKL94}.
Since they are Markovian, there are furthermore methods for synthesizing memoryless policies. Please note that 
while in principle there are finite \glspl*{BMDP} (e.g., Fig.~\ref{fig:coffee_bel_mdp1}), in general they are of infinite size~\cite{DBLP:conf/atva/BorkJKQ20} (see, e.g., Fig.~\ref{fig:coffee_bel_mdp2}). 

\begin{example}[Hot Beverage \glspl*{BMDP}]
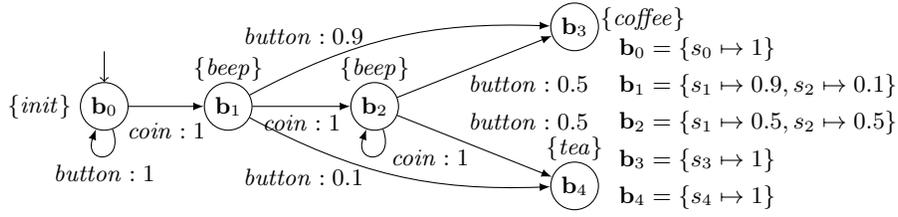
\begin{figure}[t]
    \centering
\input{figures/coffee_bel_mdp1}
    \caption{A finite \gls*{BMDP} for the \gls*{POMDP} from Fig.~\ref{fig:coffee_pomdp} and parameters $p_t = 0.1$ and $p_{tt} = p_{cc} = 0.5$. For brevity reasons, we do not show transitions from $\mathbf{b}_3$ and $\mathbf{b}_4$.}
    \label{fig:coffee_bel_mdp1}
\end{figure}
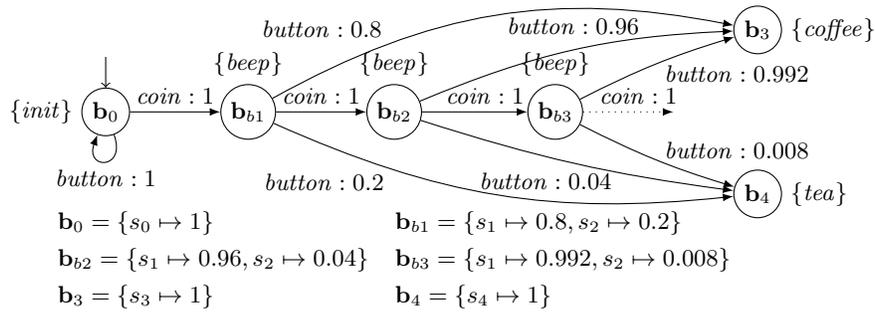
\begin{figure}[t]
    \centering
\input{figures/coffee_bel_mdp2}
    \caption{An infinite \gls*{BMDP} for the POMDP from Fig.~\ref{fig:coffee_pomdp} for parameters
    $p_t=0.2$, $p_{tt} = 0.2$, and $p_{cc} = 1$. For brevity reasons, we do not show transitions from $\mathbf{b}_3$ and $\mathbf{b}_4$.}
    \label{fig:coffee_bel_mdp2}
\end{figure}
Let us consider again the \gls*{POMDP} from Fig.~\ref{fig:coffee_pomdp} and parameters $p_t = 0.1$ and $p_{tt} = p_{cc} = 0.5$. The corresponding finite \gls*{BMDP} is shown in Fig.~\ref{fig:coffee_bel_mdp1}. Now suppose that we get a reward for observing $\mathit{tea}$, i.e., when reaching $b_4$. The \gls*{BMDP} then supports a memoryless policy where we choose the actions   $\mathit{coin}$ in $\mathbf{b}_0$ and $\mathbf{b}_1$, and $\mathit{button}$ in $\mathbf{b}_2$. That is, unless $\gamma$ is very small (like $0$) s.t. choosing $\mathit{button}$ in $\mathbf{b}_1$ would be optimal for maximizing the immediate reward. A second, infinite \gls*{BMDP} for parameters $p_t=0.2$, $p_{tt} = 0.2$ and $p_{cc} = 1$ is shown in Figure~\ref{fig:coffee_bel_mdp2}.  
\end{example}

\subsection{Learning MDPs}
We learn MDPs using the \IOALERGIA~algorithm~\cite{DBLP:journals/ml/MaoCJNLN16}. \IOALERGIA~
takes samples $\sample$, which is a multiset of traces, and 
an $\epsal$ controlling the significance level
of a statistical check as inputs and returns a deterministic labelled MDP.

The algorithm first creates an input/output frequency prefix 
tree acceptor (IOFPTA) from $\sample$, a tree where 
common prefixes of traces are merged. Every node of the tree is labeled
with an observation and every edge is labeled with input and a frequency.
The frequency denotes the multiplicity of the trace prefix in $\sample$
that corresponds to the path from the root node to the edge.
After creating a tree, \IOALERGIA~ creates an MDP by merging 
nodes that are \emph{compatible} and promoting nodes to 
MDP states that are not compatible with other states.
Initially, the root node is promoted to be the initial MDP state
and labeled \emph{red}. Then, the algorithm performs a loop 
comprising the following steps.
All immediate successors of red states are labeled \emph{blue}. A
blue node $b$ is selected and checked for compatibility with all red
states. If there is a compatible red state $r$, $b$ and $r$ are merged
and the subtree originating in $b$ is folded into the currently created
MDP. Otherwise, $b$ is labeled red, thus being promoted to an MDP state.
The loop terminates when there are only red states.

Nodes $b$ and $r$  
are compatible if their observation labels are the same and the 
probability distributions of future observations conditioned on actions
are not statistically different. The latter check is also performed 
recursively on all successors of $b$ and $r$. The statistical difference 
is based on Hoeffding bounds~\cite{10.2307/2282952}, where a parameter $\epsal$
controls significance.
A data-dependent 
$\epsal$ guarantees convergence in the limit to an MDP isomorphic 
to the canonical MDP underlying the distribution of traces.
For finite sample sizes, we can use $\epsal$ to influence the MDP size.
For more information, we refer to Mao et al~\cite{DBLP:journals/ml/MaoCJNLN16}. 

\input{figures/officeWorldFigs}

%% file: figures/coffee_pomdp.tex
\begin{tikzpicture}[node distance = 2.3cm,every node/.style={font=\footnotesize}]
   \node[initial above=,initial text=,state,inner sep=2pt,minimum size = 0.3cm](s_0){$s_0$};
   \node[state, above right= 0.4cm and 3.20cm of s_0,inner sep=2pt,minimum size = 0.3cm](s_1){$s_1$};
   \node[state, below right= 0.4cm and 3.20cm of s_0,inner sep=2pt,minimum size = 0.3cm](s_2){$s_2$};
   \node[state, right = 3.00cm of s_1,inner sep=2pt,minimum size = 0.3cm](s_3){$s_3$};
   \node[state, right = 3.00cm of s_2,inner sep=2pt,minimum size = 0.3cm](s_4){$s_4$};
   
  \path[-latex] (s_0)  edge [loop below,min distance=5mm,in=250,out=290] node[] {$\mathit{button} : 1$} (s_0)
  edge [] node[above=0.2cm] {$\mathit{coin} : 1-p_t$} (s_1)
  edge [] node[below=0.2cm] {$\mathit{coin} : p_t$} (s_2)
  (s_1)  edge [min distance=5.5mm,in=150,out=100] node[above left= -0.3cm and 0.1cm] 
{$\mathit{coin} : p_{cc}$} (s_1)
  edge [] node[above] {$\mathit{button} : 1$} (s_3)
  edge [bend left = 35] node[right] {\shortstack{$\mathit{coin} :$\\ $1-p_{cc}$}} (s_2)
  (s_2) edge node[above] {$\mathit{button} : 1$} (s_4)
      edge [bend left = 34] node[left = 0.05cm]
      {\shortstack{$\mathit{coin} :$\\ $1-p_{tt}$}} (s_1)
        (s_2)  edge [min distance=5.5mm,in=230,out=280] node[below left= -0.4cm and 0.1cm] 
{$\mathit{coin} : p_{tt}$} (s_2)
  ;
  \node[left = 0cm of s_0]{$\{\mathit{init}\}$};
  \node[above right = -0.1cm and -0.1cm of s_1]{$\{\mathit{beep}\}$};
  \node[below right = -0.1cm and 0cm of s_2]{$\{\mathit{beep}\}$};
  \node[below = -0.1cm of s_3]{$\{\mathit{coffee}\}$};
  \node[above = -0.1cm of s_4]{$\{\mathit{tea}\}$};
  \end{tikzpicture}

%% file: figures/coffee_bel_mdp1.tex
\begin{tikzpicture}[node distance = 2.3cm,every node/.style={font=\footnotesize}]
   \node[initial above=,initial text=,state,inner sep=2pt,minimum size = 0.3cm](s_0){$\mathbf{b}_0$};
   \node[state, right= 1.00cm of s_0,inner sep=2pt,minimum size = 0.3cm](s_1){$\mathbf{b}_1$};
   \node[state, right= 1.30cm of s_1,inner sep=2pt,minimum size = 0.3cm](s_2){$\mathbf{b}_2$};
   \node[state, above right = 0.6cm and 2.20cm of s_2,inner sep=2pt,minimum size = 0.3cm](s_3){$\mathbf{b}_3$};
   \node[state, below right = 0.6cm and 2.20cm of s_2,inner sep=2pt,minimum size = 0.3cm](s_4){$\mathbf{b}_4$};
   
  \path[-latex] (s_0)  
  edge [loop below,min distance=5mm,in=250,out=290] node[] {$\mathit{button} : 1$} (s_0) edge [] node[below=0.1cm] {$\mathit{coin} : 1$} (s_1)
     (s_1) 
     edge [] node[below] {$\mathit{coin} : 1$} (s_2)
     edge [bend left=15] node[above left=-0.2cm and 0.3cm] {$\mathit{button} : 0.9$} (s_3)
     edge [bend right=15] node[below left=-0.2cm and 0.3cm] {$\mathit{button} : 0.1$} (s_4)
     (s_2)  
     edge [loop below,min distance=5mm,in=250,out=290] node[below right=-0.2cm and 0.1cm] {$\mathit{coin} : 1$} (s_2)
     edge [] node[below right=0cm and -0.2cm] {$\mathit{button} : 0.5$} (s_3)
     edge [] node[above right=0.1cm and -0.2cm] {$\mathit{button} : 0.5$} (s_4)
  ;
  \node[left = 0cm of s_0]{$\{\mathit{init}\}$};
  \node[above = -0.1cm  of s_1]{$\{\mathit{beep}\}$};
  \node[above = -0.1cm of s_2]{$\{\mathit{beep}\}$};
  \node[above right = -0.4cm and 0cm of s_3]{$\{\mathit{coffee}\}$};
  \node[above = -0.1cm of s_4]{$\{\mathit{tea}\}$};
  
  \node[below right = -1.3cm and 6.5cm of s_0]{$\begin{aligned}
  \mathbf{b}_0 &= \{s_0 \mapsto 1\} \\
  \mathbf{b}_1 &= \{s_1 \mapsto 0.9, s_2\mapsto 0.1\} \\
  \mathbf{b}_2 &= \{s_1 \mapsto 0.5, s_2\mapsto 0.5\} \\
  \mathbf{b}_3 &= \{s_3 \mapsto 1\} \\
  \mathbf{b}_4 &= \{s_4 \mapsto 1\} \\
  \end{aligned}$}; 
  \end{tikzpicture}

%% file: figures/coffee_bel_mdp2.tex
\begin{tikzpicture}[node distance = 2.3cm,every node/.style={font=\footnotesize}]
   \node[initial above=,initial text=,state,inner sep=2pt,minimum size = 0.3cm](s_0){$\mathbf{b}_0$};
   \node[state, right= 1.20cm of s_0,inner sep=2pt,minimum size = 0.3cm](s_b1){$\mathbf{b}_{b1}$};
   \node[state, right= 1.20cm of s_b1,inner sep=2pt,minimum size = 0.3cm](s_b2){$\mathbf{b}_{b2}$};
   \node[state, right= 1.40cm of s_b2,inner sep=2pt,minimum size = 0.3cm](s_b3){$\mathbf{b}_{b3}$};
   \node[right = 1.2cm of s_b3](s_b4){};
   
   \node[state, above right = 0.6cm and 2.20cm of s_b3,inner sep=2pt,minimum size = 0.3cm](s_3){$\mathbf{b}_3$};
   \node[state, below right = 0.6cm and 2.20cm of s_b3,inner sep=2pt,minimum size = 0.3cm](s_4){$\mathbf{b}_4$};
   
  \path[-latex] (s_0)  
  edge [loop below,min distance=5mm,in=250,out=290] node[] {$\mathit{button} : 1$} (s_0) 
  edge [] node[above=0.0cm] {$\mathit{coin} : 1$} (s_b1)
  (s_b1)
  edge [] node[above =0.0cm] {$\mathit{coin} : 1$} (s_b2)
  edge [bend left=20] node[above left=-0.4cm and 1.4cm] {$\mathit{button} : 0.8$} (s_3)
  edge [bend right=15] node[below left=-0.4cm and 1.4cm] {$\mathit{button} : 0.2$} (s_4)
  (s_b2)
edge [] node[above right=0.0cm and -0.5cm] {$\mathit{coin} : 1$} (s_b3)
  edge [bend left=10] node[above=0.1cm] {$\mathit{button} : 0.96$} (s_3)
  edge [bend right=5] node[below left=0.0cm and -0.6cm] {$\mathit{button} : 0.04$} (s_4)
  (s_b3)
edge [dotted] node[above right=0.0cm and -0.5cm] {$\mathit{coin} : 1$} (s_b4)
  edge [bend left=5] node[below right=-0.1cm and 0cm] {$\mathit{button} : 0.992$} (s_3)
  edge [bend right=5] node[above right=-0.1cm and 0cm] {$\mathit{button} : 0.008$} (s_4)
  ;
  \node[left = 0cm of s_0]{$\{\mathit{init}\}$};
  \node[above = 0cm of s_b1]{$\{\mathit{beep}\}$};
  \node[above = 0cm of s_b2]{$\{\mathit{beep}\}$};
  \node[above = 0cm of s_b3]{$\{\mathit{beep}\}$};
  \node[right = 0cm of s_3]{$\{\mathit{coffee}\}$};
  \node[right = 0cm of s_4]{$\{\mathit{tea}\}$};
  
  \node[below = 0.8cm of s_b2]{$\begin{aligned}
  &\mathbf{b}_0 = \{s_0 \mapsto 1\} 
  &&\mathbf{b}_{b1} = \{s_1 \mapsto 0.8, s_2\mapsto 0.2\} \\
  &\mathbf{b}_{b2} = \{s_1 \mapsto 0.96, s_2\mapsto 0.04\} 
  &&\mathbf{b}_{b3} = \{s_1 \mapsto 0.992, s_2\mapsto 0.008\} \\
  &\mathbf{b}_3 = \{s_3 \mapsto 1\} 
  &&\mathbf{b}_4 = \{s_4 \mapsto 1\} 
  \end{aligned}$}; 
  \end{tikzpicture}

%% file: figures/officeWorldFigs.tex
\begin{figure}[t]
\begin{minipage}{.5\linewidth}
      \vspace{5pt}
      \centering
      \includegraphics[width=120pt]{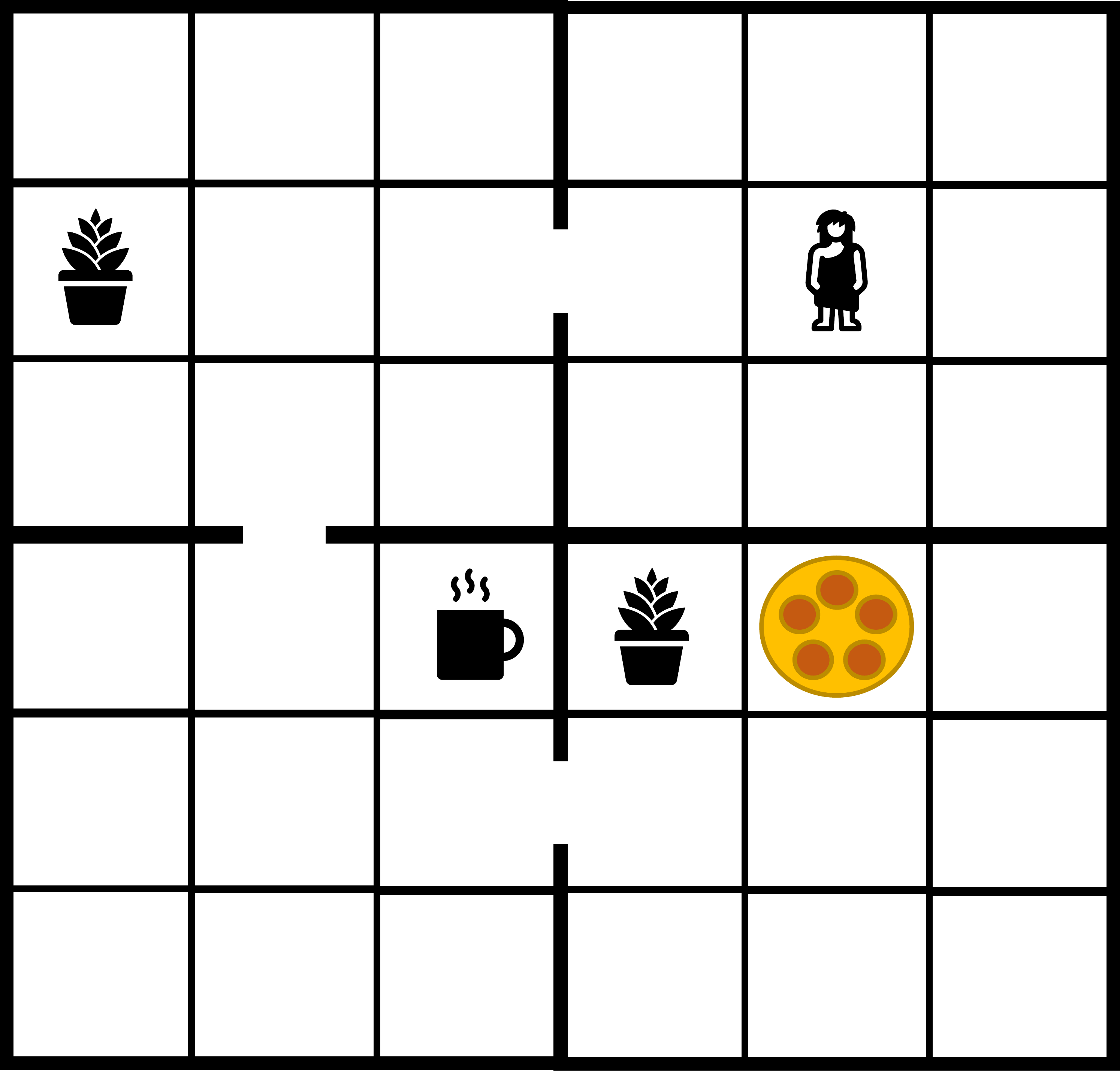}
    \end{minipage}%
    \begin{minipage}{.5\linewidth}
      \centering
      \vspace{5pt}
       \includegraphics[width=120pt]{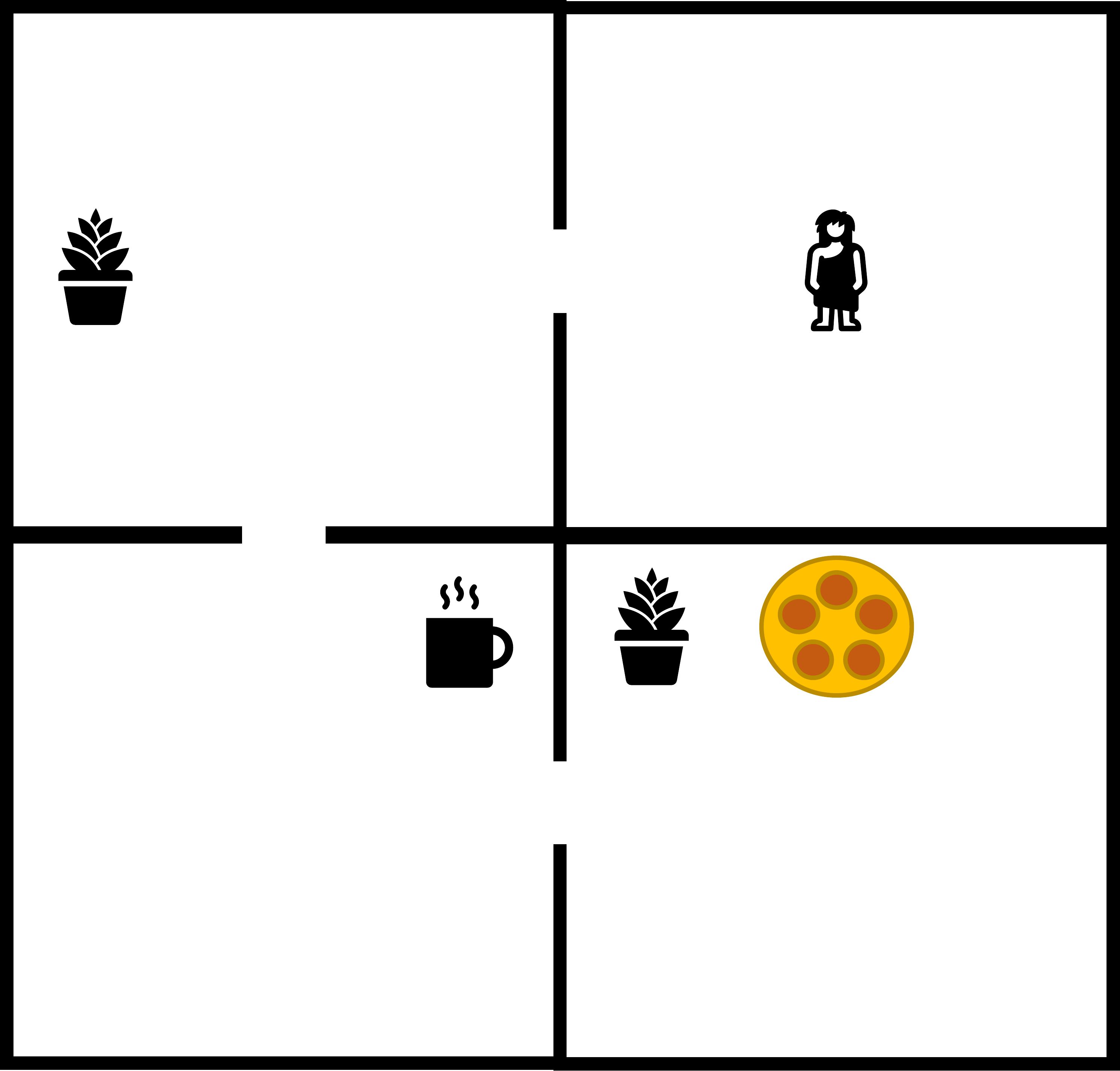}
    \end{minipage} 
\caption{Fully and partially observable \textit{OfficeWorld}. With full observations (left), the \textit{(x,y)} coordinates can be observed, otherwise (right) it is only the room's number.}
\label{fig:officeWorld}
\end{figure}

%% file: sections/method.tex
In this section, we present \poqlearning, an approach to reinforcement learning under partial observability. 
First, we describe the setting and the general intuition behind the approach. Then, we present
the state space perceived by the learning agent, followed by a presentation of the complete approach. 



\subsection{Overview}
\noindent
\emph{Setting.}
We consider reinforcement learning in partially observable environments. 
That is, we assume that the environment behaves like a POMDP, where 
we cannot observe the state directly. Moreover, 
we do not assume to have a POMDP model of the environment.
Initially we only know the available actions and as we learn, we 
learn more about the available observations and the environment dynamics
and refine our policy.

\noindent
\emph{Interface.} \label{sec:gym_interace}
We formalize the setting via an interface comprising two operations
through which the RL agent interacts with the environment: 
(1) \textbf{reset} and (2) \textbf{step}. Following the conventions 
of OpenAI gym~\cite{DBLP:journals/corr/BrockmanCPSSTZ16}, the \textbf{reset} operation resets
the environment into its initial state. The \textbf{step} operation
takes an action as input, performs the actions, which changes the
environment state, and returns the immediate reward, a Boolean flag
$\mathit{done}$, and the observation in the new state. The flag 
$\mathit{done}$ indicates whether a goal state was reached.

\noindent
\emph{Execution \& Traces.}
The agent learns in episodes, where it traverses a finite path in each episode. We want
to note again that the agent cannot see the state that it visited. 
Each executed path yields a finite reward-observation trace $rt$ 
consisting of observations, immediate rewards, and the performed 
actions. We store these reward-observation traces in a multiset $\rewsample$.

\subsection{Extended State Space}\label{sect:state_soace_extend}
The $Q$-table in Q-learning is a function $Q: S \times A \rightarrow \mathbb{R}$, where $S$ are the observable states of the environment. Since
we only observe observations from a set $Z$, we cannot use this function definition directly. As individual observations are insufficient to facilitate
learning, we extend the observation space with states of a learned MDP leading to an extended
state space. 
Suppose we are in episode $i$, 
we combine $Z$ with the states of the last labeled MDP 
$(\mathcal{M}_i, Z, O_i)$, with $\mathcal{M}_i = (S_i,{s_0}_i,A,\delta_i)$, learned via \IOALERGIA. During training, we continuously 
simulate the observation 
traces perceived by the RL agent on the learned automaton and use the visited states
from $S_i$ as additional observations. Since a learned MDP may not define
transitions for all action-observation pairs, we represent the current state 
of $\mathcal{M}_i$ as a pair $(s,d) \in S_i \times \{\top, \bot\}$, where the 
first element encodes the last visited state of $\mathcal{M}_i$ and the second
element denotes whether the simulation encountered an undefined transition.
To work with these state pairs, we define two functions, where for $a \in A$ and $o \in O$:
\begin{align*}
    \mathit{resetToInitial}() &= ({s_0}_i, \top) \\
    \mathit{stepTo}((s,d), a, o) &= \begin{cases}
    (s', \top) & \text{ if } d = \top \land \delta_i(s,a)(s')>0 \land O_i(s') = o \\
    (s, \bot) & \text{ otherwise }
    \end{cases}
\end{align*}

The extended state space uses these state pairs, i.e.,  
the $Q$-function is defined as $Q: S^e_i \times A \rightarrow \mathbb{R}$
with $S^e_i =  O \times S_i \times \{\top,\bot\}$.  
Due to $\mathcal{M}_i$ being deterministic, $s'$ in the above
definition is either uniquely
defined or not defined at all, denoted by $\bot$. Once we reach 
undefined behavior, we remember the last visited state and leave it unchanged.
The intuition is that when behavior is encountered after reaching some $(s, \bot)$
is important for RL performance, due to achieving high reward, this will be
reflected in updates of the $Q$-function. As a result, 
learning will be directed towards $s$. This leads to more sampling in the
vicinity of $s$ s.t.\ subsequently learned MDPs are more accurate in 
this region. Consequently, previously undefined behavior will eventually
become defined in the learned MDP.


\subsection{Partially Observable Q-Learning}
\input{figures/mainAlgorithm}
We apply tabular, $\epsilon$-greedy Q-learning~\cite{Watkins1992} 
combined with MDP learning.
Deterministic labeled MDPs learned by \IOALERGIA~provide the Q-learning
agent with additional information in order to make the 
learning problem Markovian despite partial observability. 

We regularly learn new MDPs via \IOALERGIA~from the growing sample
of reward-observation traces, where we discard the rewards,
so that at each episode $i$ there is an
approximate MDP $\mathcal{M}_i$ with states $S_i$. To take information 
from $\mathcal{M}_i$ into account during RL, we extend the Q-table with
observations corresponding to the states $S_i$.
At every step performing action $a$ and observing $o$ during RL, 
we simulate the step in $\mathcal{M}_i$. This yields a unique
state in $S_i$ due to $\mathcal{M}_i$ being deterministic, 
which we feed to the RL agent as an additional observation.


We actually perform two stages of learning.
First, 
we perform Q-learning while regularly updating $\mathcal{M}_i$. 
In the second stage, we fix the final MDP $\mathcal{M}_i$, referred to \emph{freezing} below, 
and perform Q-learning without learning new MDPs with \IOALERGIA. We term the resulting learning approach \poqlearning.

Algorithm \ref{alg:poql:training} implements this learning approach, i.e., training of a \poqlearning\  agent. For a more detailed view of the training algorithm and agent parameterization, we point an interested reader to the implementation~\footnote{\url{https://anonymous.4open.science/r/Q-learning-under-Partial-Observability-4BEC}}. 

Algorithm \ref{alg:poql:training} assumes that the \poqlearning~agent interacts with the environment \textit{env} as described in Section.~\ref{sec:gym_interace}. 
The parameter \textit{maxEp} defines the maximum number of training episodes. The other parameter \textit{updateInterval} defines how often the agent recomputes the model, thus extending the state space perceived by the learning agent and the Q-table.

Lines \ref{algline:poql:init_start}-\ref{algline:poql:init_end} initialize the 
extended state space and set the actions of the agent to those of the environment. 
For the state-space initialization, we assume an initial approximate MDP to be given.
In our implementation, we learn such an MDP from a small number of randomly generated traces.
Alternatively, the extended state space \textit{$S_{E}$} can be initialized to the observation space of the environment. It will in any case be extended as the algorithm progresses. Line \ref{algline:poql:initial_q_table} initializes the Q-table with the initial observation space and action space. 

Training progresses until the maximum number of episodes is reached. In the implementation, we have added an early stopping criterion to end the training as soon as the agent achieves satisfactory performance on a predefined number of test episodes. Lines \ref{algline:poql:ep_begin}-\ref{algline:poql:save_obs} show the steps taken in a single training episode. At the beginning of each episode, the environment and the agent's internal state are reset to their initial states (Lines \ref{algline:poql:env_reset} and \ref{algline:poql:reset_automaton}). Until an episode terminates, either by reaching a goal or exceeding the maximum number of allowed steps, the \poqlearning~agent selects an action using an $\epsilon$-greedy policy and executes it in the environment (Lines \ref{algline:poql:action_selection}-\ref{algline:poql:env_output}). Based on the selected action \textit{act} and received observation \textit{newObs}, the agent updates its current model state by tracing the pair \textit{(act, newObs)} in the learned MDP (Line \ref{algline:poql:step_automaton}). 
After performing a step, the agent updates the values in the Q-tables based on observations and received reward. Alg.~\ref{alg:poql:q_values_update} describes the process of updating Q-values. It follows the same procedure as in standard Q-learning with the notable difference of using a state space extended with learned MDP states instead of the observation space of the environment. The extended state space is discussed in more detail in Sect.~\ref{sect:state_soace_extend}

In Line \ref{algline:poql:freeze_automaton}, we check whether the automaton should be \emph{frozen}. Freezing of the automaton prevents further updates of the model and extensions of the state space. This way once the automaton is frozen, the Q-table will continue to be optimized with respect to the current extended state space. Automaton freezing operates under the assumption that once a model is computed that is ``good enough'', computing a new model in the next update interval is unnecessary and might even be detrimental to the performance of the agent (in the short term). 

If the automaton freezing is not enabled or its episode threshold has not yet been reached, 
we proceed with the update of the model and the Q-table (Lines \ref{algline:poql:run_alergia}-\ref{algline:poql:extend_q_table}).
This update happens every \textit{updateInterval} episodes. \IOALERGIA~computes a new model that conforms to the sample $\rewsample$ with rewards discarded. In Line \ref{algline:poql:extend_state_space} we extend the state space with state identifiers
of the learned model. After that, we recompute the Q-table by initializing it with the extended state space and action space (Line ~\ref{algline:poql:extend_q_table}). To recompute the values in the extended Q-table we perform an experience replay~\cite{DBLP:conf/aaai/GaonB20} with all traces in $\rewsample$ (Lines \ref{algline:poql:resample_start}-\ref{algline:poql:resample_end}).

\input{figures/q_tables}

\subsubsection{Demonstrating Example.}
We use a simple \textit{OfficeWorld} example to demonstrate state extension. 
In this example,
an agent selects one of four actions: \{\textit{up, down, left, right}\} to move into the given direction. The agent may also slip into a different direction with a location-specific probability.

Whereas~\cite{DBLP:conf/aaai/NeiderGGT0021,DBLP:conf/icml/IcarteKVM18} use \textit{OfficeWorld} in a non-Markovian reward setting under full observability, we modify the \textit{OfficeWorld} layout as shown in Fig.~\ref{fig:officeWorld} to introduce partial observability. 
On the right-hand side of Fig.~\ref{fig:officeWorld}, abstraction is applied over the state space. The reinforcement learning agent can only observe in which room he is located, but not the $x$ and $y$ coordinates, which would truly identify a Markov state. 
Note that each observation, e.g. \textit{Room1}, is 
shared by nine different POMDP states identified by their $x$, $y$ coordinates, each with different future and stochastic behavior.

Table~\ref{table:initial_q_table} shows a Q-table obtained from observations only. 
The Q-values indicate that the Q-learning agent is unable to find optimal actions due to partial observability.
Table~\ref{table:extanded_q_table} shows an extended Q-table, 
but we do not include the definedness flag from $\{\top,\bot\}$ for brevity. Each observation is extended with a learned MDP state as discussed in Sect.~\ref{sect:state_soace_extend}. We observe that each \textit{(observation, state)} pair approximates the underlying BMDP to such an extend that every such pair has a clear optimal action defined by the Q-values. For example, in states \textit{(Room1, s0)} and \textit{(Room1, s1)} the agent needs to perform a \textit{left} action, whereas in state \textit{(Room1, s3)} the agent needs to move \textit{up}.
We can also observe that the Q-values of the state \textit{(Room2, s0)} are set to zero. This results from \textit{s0} being unreachable 
while observing \textit{Room2}.



\begin{algorithm}[t!]
\begin{algorithmic}[1]
\footnotesize
\Require \poqlearning~\textit{agent}, environment \textit{state}, \textit{reward}, reached environment \textit{newState} 
\State $\mathit{extendedNewState} \gets \mathit{agent.model.stepTo(action, newState)}$
\State $\mathit{oldValue} \gets agent.Q(extendedState, action)$
\State $\mathit{maxNextStateValue} \gets max(agent.Q(extendedNewState))$
\State $\mathit{agent.Q(extendedState, action)} \gets (1-\alpha) * oldValue + \alpha * (reward + \gamma * maxNextStateValue)$

\end{algorithmic}
\caption{Algorithm implementing update of Q-values of the agent.}
\label{alg:poql:q_values_update}
\end{algorithm}

\subsection{Correctness}
\poqlearning~learns an optimal policy in 
the limit, when the number of episodes tends to 
infinity, if the BMDP of the POMDP environment is finite.

This property follows from the convergence of 
\IOALERGIA~and Q-learning. We sample traces 
from a POMDP that, by assumption, is  
equivalent to a finite BMDP. The BMDP 
itself is a deterministic labeled MDP. \IOALERGIA~in the 
limit learns an MDP isomorphic to the canonical
deterministic labeled MDP producing the traces
when every action always has a non-zero 
probability to be executed, as has been shown by Mao et al.~\cite{DBLP:journals/ml/MaoCJNLN16}. 
That is, every pair of belief-state and action would be 
explored infinitely often in the limit.

This also ensures convergence of Q-learning in a Markovian environment~\cite{Watkins1992}. 
The environment  
is Markovian once we learned the BMDP and add its current
state to the observations of the Q-learning agent. Hence, we
will learn an optimal policy for the BMDP and thus the POMDP in the limit.

When the BMDP is not finite and for finite sample sizes, we learn
an approximation of the BMDP. 
For instance, in the example shown in Fig.~\ref{fig:coffee_bel_mdp2}, we might learn the three
belief states labeled with $\mathit{beep}$, but beyond 
that the probability of observing $\mathit{tea}$ is  likely too
small to detect additional states. 
As we will demonstrate in 
the evaluation, such approximate MDPs encode sufficient information to aid reinforcement
learning. The learned automaton and its states can be thought of as providing 
memory to the reinforcement learning agent. 




%% file: figures/mainAlgorithm.tex
\begin{algorithm}[t!]
\begin{algorithmic}[1]
\footnotesize
\Require reinforcement learning environment  $env$, fully configured partially observable agent $\mathit{agent}$, update interval $\mathit{updateInterval}$, number of training episodes $maxEp$
\Ensure trained $\mathit{agent}$ implementing the policy for the $env$ 
\State  $\mathit{agent.S_{E}} \gets \mathit{env.O}\times agent.model.states \times \{\top, \bot\}$ \Comment{Init.\ extended state space} \label{algline:poql:init_start}
\State  $\mathit{agent.A} \gets \mathit{env.A}$ \Comment{Get action state space from $env$} \label{algline:poql:init_end}
\State $\mathit{agent.Q(s,a) = 0, \forall s \in S_{E}, a \in A}$ \Comment{Initialize the Q-table} \label{algline:poql:initial_q_table}
\State $\rewsample$ $\gets \{\}$ \Comment{Multiset of traces}
\For{$trainingEpisode \gets 0 \textbf{ to }maxExp$} \label{algline:poql:ep_begin}
\State $initialObs, initialRew \gets \mathit{env.reset()}$ \label{algline:poql:env_reset}
\State $\mathit{rt} \gets \langle initialObs, initialRew \rangle$ \Comment{Trace of a single episode}
\State $\mathit{agentState} \gets \mathit{agent.model.resetToInitial()}$ \label{algline:poql:reset_automaton}
\State $epDone \gets False$
\While{$\mathbf{not}~epDone$}
\State \Comment{Select an action using $\epsilon$-greedy policy and extended state space}
\State $\mathit{act} \gets \mathit{agent.getAction(agent.state)}$  \label{algline:poql:action_selection}
\State \Comment{Record all observed (state, action,reward, newState) pairs}  
\State $obs, reward, done, newObs \gets env.step(act)$ \label{algline:poql:env_output}
\State $\mathit{agentState} \gets \mathit{agent.model.stepTo(agentState,act, newObs)}$ \label{algline:poql:step_automaton}
\State $\mathit{updateQValues(agent, obs, act, reward, newObs)}$ \label{algline:poql:update_q_values}
\State $\mathit{rt} \gets \mathit{rt} \cdot \langle act, reward, newObs \rangle$ \label{algline:poql:save_obs}
\EndWhile
\State $\rewsample \gets \rewsample \uplus \mathit{rt}$ \label{algline:poql:extend_RT}

\If{$\mathit{trainingEpisode}~ \geq \mathit{agent.freezeAutomaton}$} \Comment{Freeze automaton} \label{algline:poql:freeze_automaton}
\State \textbf{continue}
\EndIf

\If{$\mathit{trainingEpisode}~\mathbf{mod}~\mathit{updateInterval}~=~0$}
\State $\mathit{agent.model} \gets \mathit{runIOAlergia(\rewsample)}$ \Comment{Learn the new environment model} \label{algline:poql:run_alergia}
\State $\mathit{agent.S_{E}} \gets \mathit{agent.S_{Init}} \times  \mathit{agent.model.states}\times \{\top, \bot\}$ \label{algline:poql:extend_state_space}
\State $\mathit{agent.Q(s,a) = 0, \forall s \in S_{E}, a \in A}$  \Comment{Reinitialize the extended Q-table} \label{algline:poql:extend_q_table}
\For{$\textit{episode} \in \rewsample$} \label{algline:poql:resample_start}
\State $\mathit{agentState} \gets \textit{agent.model.resetToInitial()}$
\For{$\textit{obs, action, reward, newObs} \in \textit{episode}$} 
\State $\mathit{agentState} \gets \mathit{agent.model.stepTo(\mathit{agentState},act, newObs)}$
\State $\mathit{updateQValues(agent, obs, action, reward, newObs)}$ \label{algline:poql:resample_end}
\EndFor
\EndFor
\EndIf

\EndFor
\Return $\mathit{agent}$

\end{algorithmic}
\caption{Algorithm implementing \poqlearning}
\label{alg:poql:training}
\end{algorithm}

%% file: figures/q_tables.tex
\begin{wraptable}{R}{6cm}

\caption{Non-extended Q-table.}
\label{table:initial_q_table}
\centering
\begin{tabular}{|l|l|l|l|l|}
    \hline
    State\textbackslash{}Action & Up    & Down  & Left  & Right \\ \hline
    Room1                       & -0.35 & -0.35 & -0.45 & -0.39 \\ \hline
    Room2                       & -0.67 & -0.67 & -0.32 & -0.66 \\ \hline
    Room3                       & 4.68  & 4.65  & 5.21  & 5.25  \\ \hline
    Room4                       & 24.72 & 24.45 & 23.26 & 24.79 \\ \hline
\end{tabular}%

\centering
\caption{Extended Q-table.}
\label{table:extanded_q_table}

    \begin{tabular}{|lllll|}
    \hline
    \multicolumn{1}{|l|}{State\textbackslash{}Action} & \multicolumn{1}{l|}{Up}    & \multicolumn{1}{l|}{Down}  & \multicolumn{1}{l|}{Left} & Right \\ \hline
    \multicolumn{1}{|l|}{(Room1, s0)}                 & \multicolumn{1}{l|}{-0.35} & \multicolumn{1}{l|}{-0.35} & \multicolumn{1}{l|}{0.75} & -0.39 \\ \hline
    \multicolumn{1}{|l|}{(Room1, s1)}                 & \multicolumn{1}{l|}{-0.33} & \multicolumn{1}{l|}{-0.35} & \multicolumn{1}{l|}{0.90} & -0.36 \\ \hline
    \multicolumn{1}{|l|}{(Room1, s2)}                 & \multicolumn{1}{l|}{-0.31} & \multicolumn{1}{l|}{-0.31} & \multicolumn{1}{l|}{1.06} & -0.30 \\ \hline
    \multicolumn{1}{|l|}{(Room1, s3)}                 & \multicolumn{1}{l|}{0.4} & \multicolumn{1}{l|}{-0.41} & \multicolumn{1}{l|}{0.06} & -0.35 \\ \hline
    \multicolumn{5}{|l|}{...}                                                                                                                       \\ \hline
    \multicolumn{1}{|l|}{(Room2, s0)}                 & \multicolumn{1}{l|}{0}     & \multicolumn{1}{l|}{0}     & \multicolumn{1}{l|}{0}    & 0     \\ \hline
    \multicolumn{1}{|l|}{(Room2, s5)}                 & \multicolumn{1}{l|}{-0.67} & \multicolumn{1}{l|}{1.23}  & \multicolumn{1}{l|}{1.07} & -0.66 \\ \hline
    \multicolumn{1}{|l|}{(Room2, s6)}                 & \multicolumn{1}{l|}{-0.67} & \multicolumn{1}{l|}{1.42}  & \multicolumn{1}{l|}{0.4}  & -0.66 \\ \hline
    \multicolumn{5}{|l|}{...}                                                                                                                       \\ \hline
    \multicolumn{1}{|l|}{(Room4, s10)}                & \multicolumn{1}{l|}{97.3}  & \multicolumn{1}{l|}{74.1}  & \multicolumn{1}{l|}{78.2} & 93.8  \\ \hline
    \multicolumn{1}{|l|}{(Room4, s11)}                & \multicolumn{1}{l|}{97.9}  & \multicolumn{1}{l|}{74.7}  & \multicolumn{1}{l|}{78.8} & 92.1  \\ \hline
    \multicolumn{1}{|l|}{(Room5, s12)}                & \multicolumn{1}{l|}{98.3}  & \multicolumn{1}{l|}{75.2}  & \multicolumn{1}{l|}{79.2} & 95.3  \\ \hline
\multicolumn{5}{|l|}{...}                                                       
                                                                \\ \hline
\end{tabular}%
\end{wraptable}

%% file: sections/eval.tex
To evaluate the proposed method we have implemented \poqlearning~in Python. The implementation uses \textsc{AALpy}'s~\cite{muskardin_aalpy:_2021} \IOALERGIA~implementation and it interfaces OpenAI \textit{gym}~\cite{DBLP:journals/corr/BrockmanCPSSTZ16}. The implementation can be used on all \textit{gym} environments with discrete action and observation space. We have evaluated \poqlearning\ by comparing its performance on four partially observable environments with multiple state-of-the-art RL algorithms implemented in OpenAI's stable-baselines~\cite{stable-baselines}, considering: 
\textit{non-recurrent policies with a stacked history of observations} and \textit{LSTM-based policies}. 

\textit{Stacked history of observations.} In a first set of experiments, we have compared \poqlearning~with DQN~\cite{DBLP:journals/corr/MnihKSGAWR13}, A2C~\cite{DBLP:conf/icml/MnihBMGLHSK16}, and ACKTR~\cite{DBLP:conf/nips/WuMGLB17}. To aid those algorithms to cope with partial observablility, we encoded the history of observations as a stacked frame. Stacked observation frames encode an observation history by using the last $n$ observations observed during training and evaluation. By using stacked observations, we extend the observation space from initially $i$ observations to $(i+1)^n$ observations \footnote{+1 due to the padding 
observation present in the first $n$ steps of each training episode}. For all experiments we have set the size of stacked frames to 5. A similar approach was, e.g., used in~\cite{DBLP:journals/corr/MnihKSGAWR13} as a method to encode movement in ATARI games. 

\textit{LSTM-based policies.} Deep-recurrent Q-learning~\cite{DBLP:conf/aaaifs/HausknechtS15} has been used to solve ATARI games without stacking the history of observations. Hausknecht and Stone~\cite{DBLP:conf/aaaifs/HausknechtS15} show that recurrence is a viable alternative to frame stacking, and while no significant advantages were noticed during training of the agent, LSTM-based policies were more adaptable in the evaluation phase in the presence of previously unseen observations. 

\textit{Setup.}
All experiments were conducted on a laptop with an Intel\textsuperscript{\textregistered} Core\texttrademark 
i7-11800H at 2.3 GHz, 32 GB RAM, and an NVIDIA RTX\texttrademark 3050 Ti graphics card using Python3.6. For all experiments we have set the maximum number of training episodes to 30,000. A training episode ends if an agent reaches a goal or the maximum number of steps is exceeded. The training performance was periodically evaluated and training was halted when reaching satisfactory performance. 

Table~\ref{table:eval_results} summarizes the results of the experiments. There are columns for each partially observable environment, where the first shows the average number of actions required to reach a dedicated goal with the best policy
found by RL, and the second column shows the number of training episodes needed to learn
a policy. The symbol \ding{55} denotes that no policy was found that reaches
the goal within the allotted maximum number of steps. 
The rows correspond to: the optimal policy, the policy found by
\poqlearning, the three RL approaches with stacked observations, and the three
approaches with LSTM-based policies.
All experiments were repeated multiple times and we chose the best training run as a representative for each approach. As the agent performance was evaluated on 100 episodes, the average number of steps to reach a goal was rounded to the closest integer. In the remainder of this section, we will explain the partially observable environments on which the agents were trained and discuss the obtained results.

    The \textit{OfficeWorld} domain is depicted on the right of Fig.~\ref{fig:officeWorld}. \poqlearning{} was able to find an optimal policy in this environment, but with a higher total number of training episodes compared to LSTM-based approaches. Stacked-frame based approaches also performed well, but were not able to find an optimal policy. This environment was solvable by all approaches despite its partial observability as each room has two actions which when executed repeatedly will lead the agent into the next room (e.g., in Room2 the agent needs to repeatedly perform down and right actions).
    
    \textit{ConfusingOfficeWorld} found on the left-hand side of Fig.~\ref{fig:evalFigues} is a variation of the  \textit{OfficeWorld}. \textit{ConfusingOfficeWorld} is harder to solve as 
    the agent receives the same observations in the upper right and the lower left rooms, likewise in the upper left and the lower right rooms. 
    The rooms labeled with \textit{Room1} and have two sets of opposite actions that need to be taken, depending on the actual agent location. The same
    holds for \textit{Room2}.
    \poqlearning~was able to find a solution for this world in 16 thousand episodes, while other approaches failed due to insufficient state differentiation.
    
    \textit{GravityDomain} was inspired by the environment discussed in~\cite{DBLP:conf/icml/IcarteKVM18}. In \textit{GravityDomain}, gravity will pull the agent down in each state with 50\% probability. By reaching a toggle indicated by a blue switch in Fig.~\ref{fig:gravityDomain}, gravity is turned off and 
    the environment becomes deterministic. We observed that both stacked-frame and LSTM-based approaches learned a policy in which they repeatedly performed the \textit{up} action, thus reaching the goal in only 50\% of the test episodes
    within the maximum number of $100$ steps. \poqlearning~was able to learn an optimal strategy in which it first reached the blue toggle and then proceeded to the goal, depicted by a cookie. Note that the approach presented in~\cite{DBLP:conf/icml/IcarteKVM18} is generally not able to solve the \textit{GravityDomain}.
    
    \textit{ThinMaze} is depicted on the left-hand side of Fig.~\ref{fig:evalFigues}. In \textit{ThinMaze}, the only observations are 
    ``cookie'' and ``wall'', which signals that the agent performs an action that is blocked by a wall. Due to the lack of observations both stacked-frame and LSTM-based approaches failed to find a solution to \textit{ThinMaze}. \poqlearning~ was able to find a non-optimal solution in 27 thousand training episodes. This is due to the fact that \IOALERGIA~requires a high number of traces to approximate the underlying belief-MDP with sufficient accuracy. 

\emph{Runtime.} We only briefly comment on runtime, considering \emph{OfficeWorld} for a fair comparison, as all approaches found a decent policy. Other results might be skewed
due to different training lengths. Stacked-frame DQN, A2C, and ACKTR, require $312s$, 
 $373s$, and $48s$, respectively. Stopping after only $1k$ episodes, the LSTM-backed ACER, A2C, and ACKTR take $51s$, $74s$, and $80s$, respectively. Our approach is considerably faster, finishing after about $3s$. The reason is that we apply tabular Q-learning and \IOALERGIA~adds very little runtime overhead. The automata-learning technique has cubic worst-case runtime in the sample size, but has been reported to have linear runtime in practice~\cite{DBLP:conf/icgi/CarrascoO94}.

\input{figures/gravityFigs}

\input{figures/evaluationWorldsFigs}

\input{figures/eval_table}

%% file: figures/gravityFigs.tex
\begin{wrapfigure}{R}{0.45\textwidth}
\begin{minipage}{.5\linewidth}
      \vspace{5pt}
      \label{fig:gravity}
      \raggedright
      \includegraphics[width=75pt]{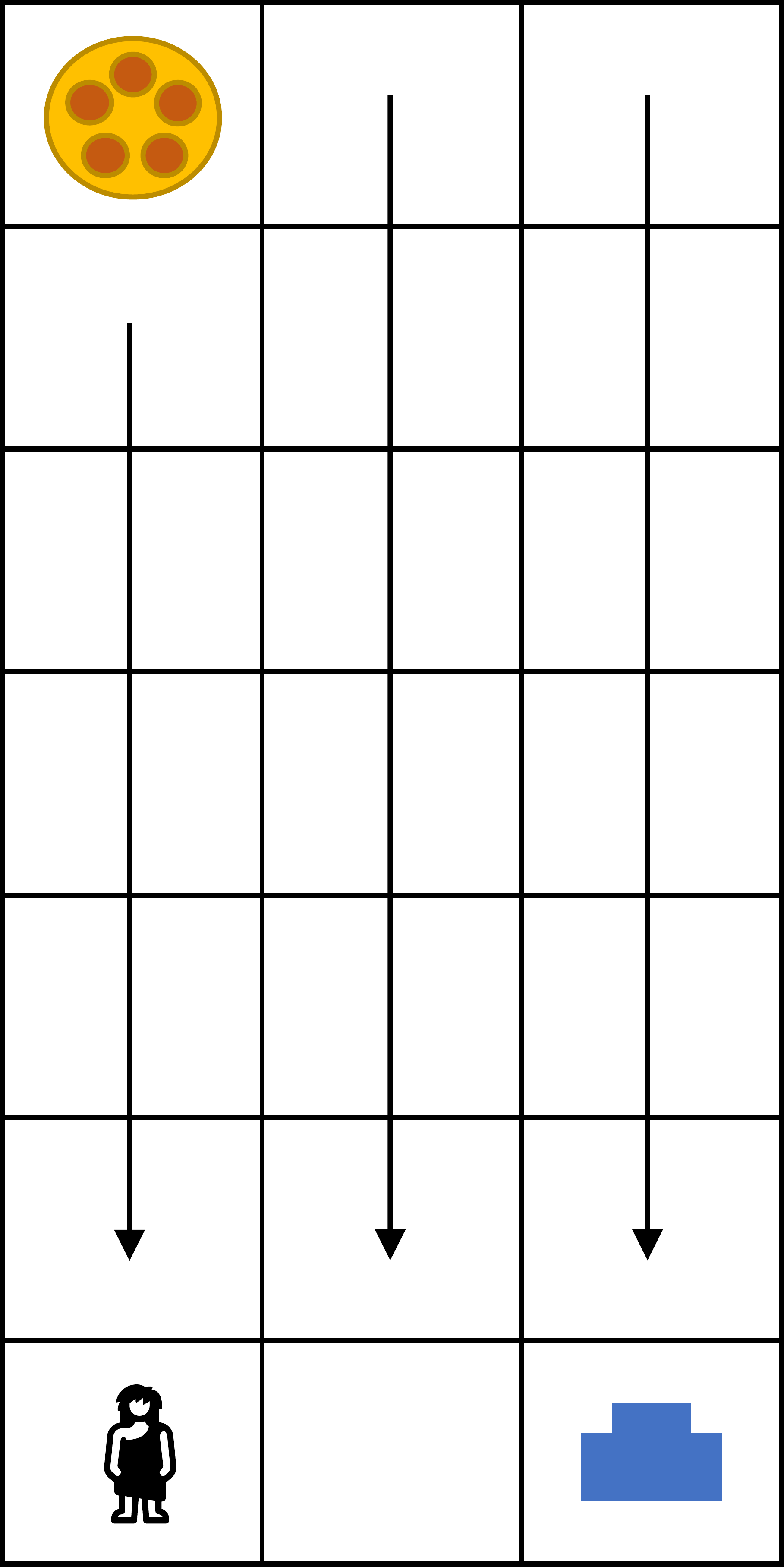}
    \end{minipage}%
    \begin{minipage}{.5\linewidth}
      \raggedleft
      \vspace{ 5pt}
      \label{fig:gravityAbstracted}
       \includegraphics[width=75pt]{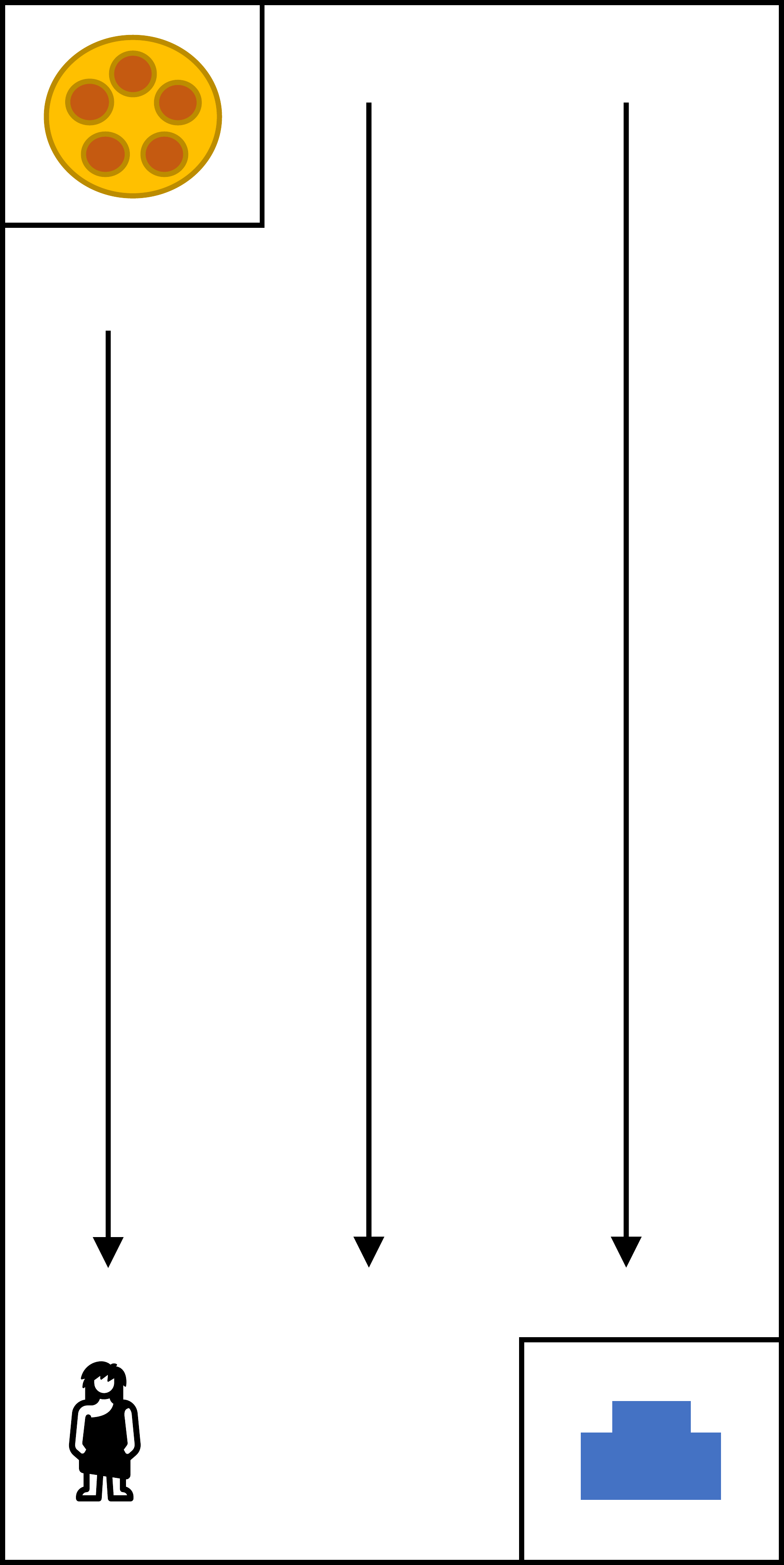}
    \end{minipage} 
\caption{Fully and partially observable gravity domain. Once the button in the lower right corner is reached, gravity is turned off. 
Please note that for conciseness, we show a width of three, while we used six states in our evaluation.
}
\label{fig:gravityDomain}
\end{wrapfigure}


%% file: figures/evaluationWorldsFigs.tex
\begin{figure}[t]
\begin{minipage}{.5\linewidth}
      \vspace{5pt}
      \centering
      \includegraphics[width=120pt]{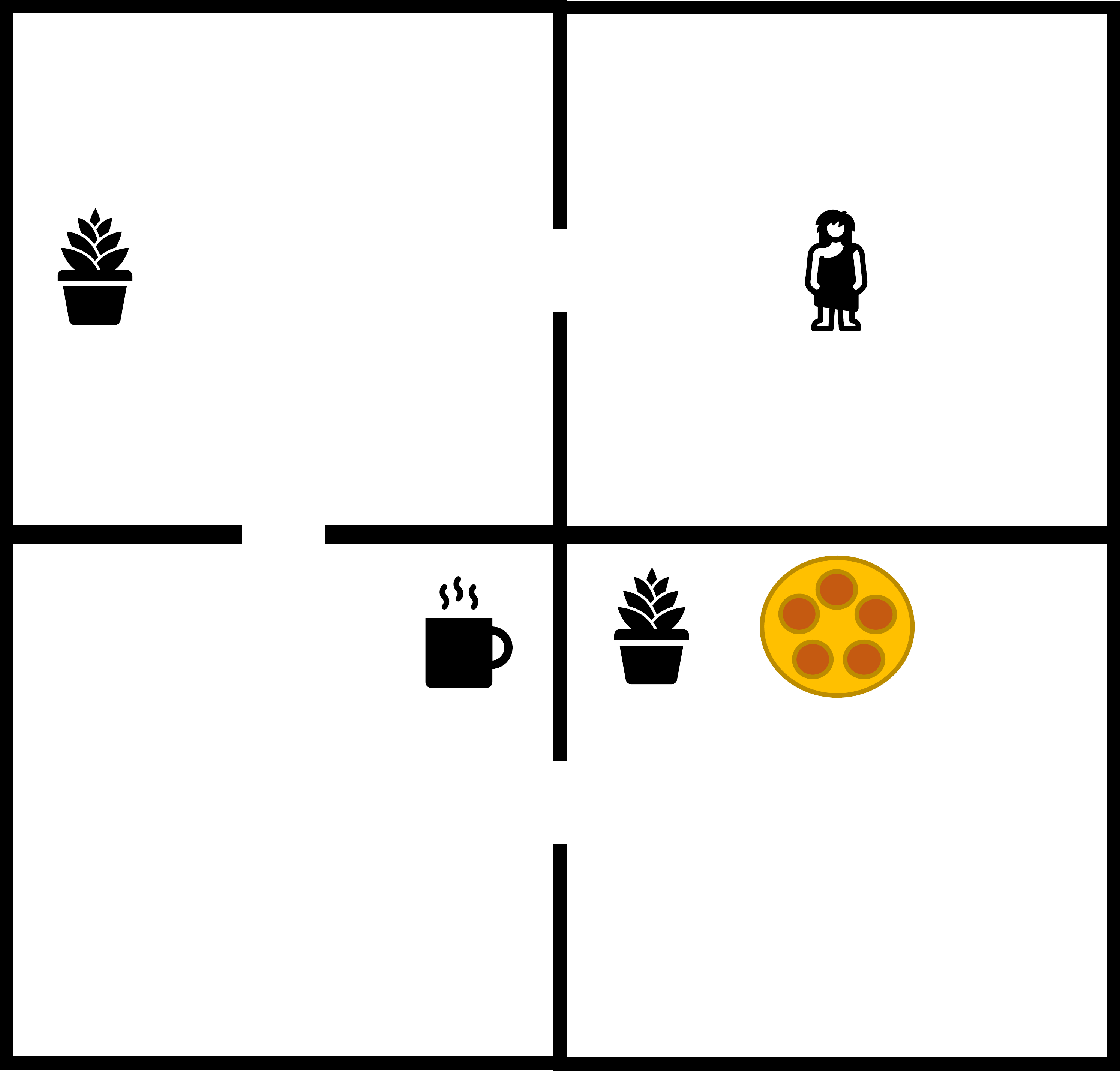}
    
    \end{minipage}%
    \begin{minipage}{.5\linewidth}
      \centering
      \vspace{5pt}
       \includegraphics[width=113pt]{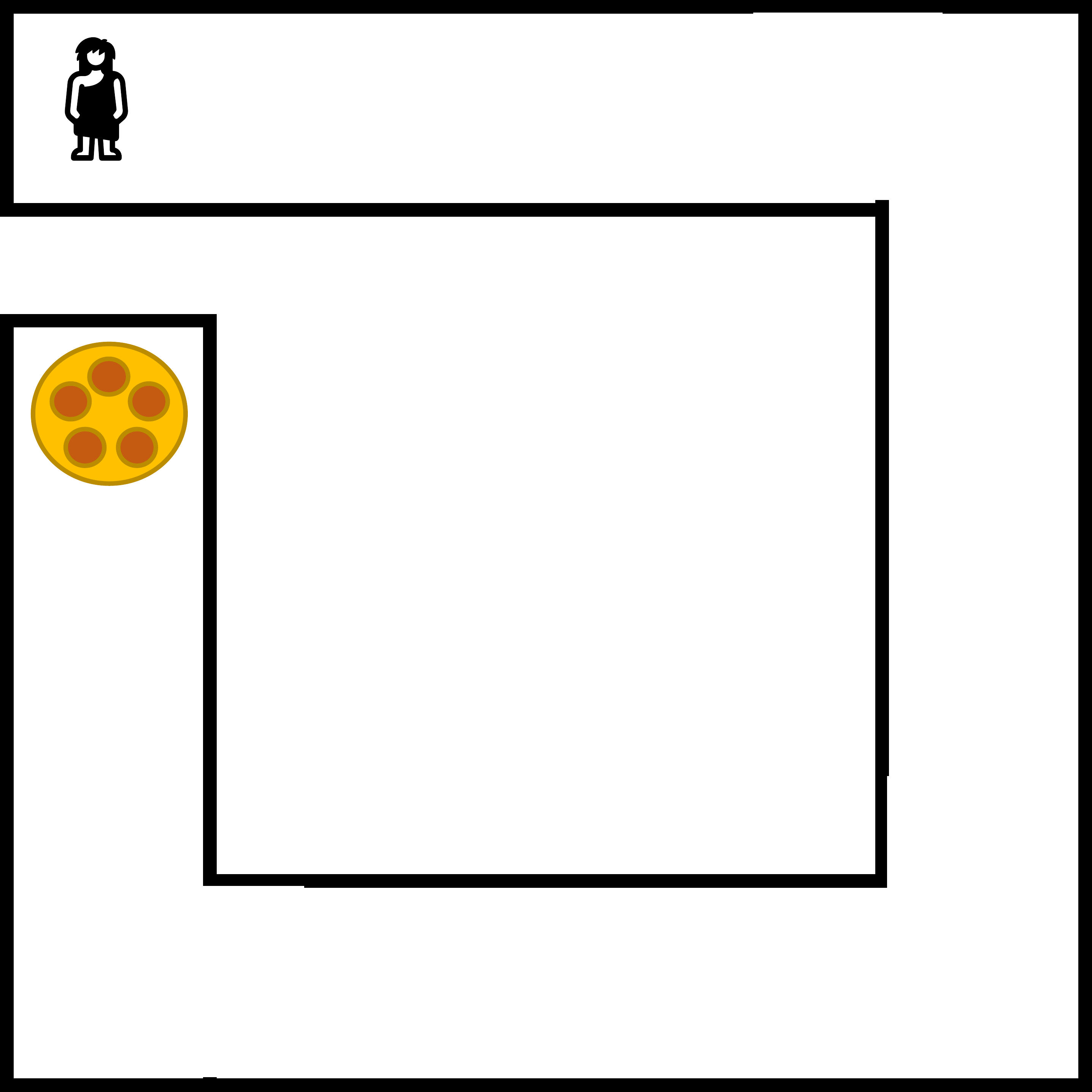}
    \end{minipage} 

\caption{\textit{ConfusingOfficeWorld} (left) and partially observable \textit{ThinMaze} (right).}
\label{fig:evalFigues}
\end{figure}

%% file: figures/eval_table.tex
\begin{table}[t]
\caption{Representative evaluation results.}
\label{table:eval_results}
\resizebox{\textwidth}{!}{%
\begin{tabular}{cc|cc|cc|cc|cc}
\multicolumn{1}{l}{}                                                                                & \multicolumn{1}{l|}{} & \multicolumn{2}{c|}{OfficeWorld}                                                                                                                                      & \multicolumn{2}{c|}{\begin{tabular}[c]{@{}c@{}}Confusing\\ OfficeWorld\end{tabular}}                                                                                  & \multicolumn{2}{c|}{GravityDomain}                                                                                                                                    & \multicolumn{2}{c}{ThinMaze}                                                                                                                                         \\ \cline{3-10} 
\multicolumn{2}{c|}{Algorithm}                                                                                              & \multicolumn{1}{l|}{\begin{tabular}[c]{@{}l@{}}\# Steps to\\ Goal\end{tabular}} & \multicolumn{1}{l|}{\begin{tabular}[c]{@{}l@{}}\# Training\\ Episodes\end{tabular}} & \multicolumn{1}{l|}{\begin{tabular}[c]{@{}l@{}}\# Steps to\\ Goal\end{tabular}} & \multicolumn{1}{l|}{\begin{tabular}[c]{@{}l@{}}\# Training\\ Episodes\end{tabular}} & \multicolumn{1}{l|}{\begin{tabular}[c]{@{}l@{}}\# Steps to\\ Goal\end{tabular}} & \multicolumn{1}{l|}{\begin{tabular}[c]{@{}l@{}}\# Training\\ Episodes\end{tabular}} & \multicolumn{1}{l|}{\begin{tabular}[c]{@{}l@{}}\# Steps to\\ Goal\end{tabular}} & \multicolumn{1}{l}{\begin{tabular}[c]{@{}l@{}}\# Training\\ Episodes\end{tabular}} \\ \hline
\multicolumn{2}{c|}{Optimal Solution}                                                                                       & \multicolumn{1}{c|}{12}                                                         & -                                                                                   & \multicolumn{1}{c|}{12}                                                         & -                                                                                   & \multicolumn{1}{c|}{18}                                                         & -                                                                                   & \multicolumn{1}{c|}{20}                                                         & -                                                                                  \\ \hline
\multicolumn{2}{c|}{\poqlearning}                                                                                           & \multicolumn{1}{c|}{12}                                                         & 3k                                                                                  & \multicolumn{1}{c|}{18}                                                         & 16k                                                                                 & \multicolumn{1}{c|}{18}                                                         & 2k                                                                                  & \multicolumn{1}{c|}{32}                                                         & 27k                                                                                \\ \hline
\multicolumn{1}{c|}{\multirow{3}{*}{\begin{tabular}[c]{@{}c@{}}Stacked\\ observation\end{tabular}}} & DQN                   & \multicolumn{1}{c|}{17}                                                         & 2k                                                                                  & \multicolumn{1}{c|}{\ding{55}}                                                      & 30k                                                                                 & \multicolumn{1}{c|}{75}                                                         & 30k                                                                                 & \multicolumn{1}{c|}{\ding{55}}                                                      & 30k                                                                                \\ \cline{2-10} 
\multicolumn{1}{c|}{}                                                                               & A2C                   & \multicolumn{1}{c|}{24}                                                         & 12k                                                                                 & \multicolumn{1}{c|}{\ding{55}}                                                      & 30k                                                                                 & \multicolumn{1}{c|}{75}                                                         & 30k                                                                                 & \multicolumn{1}{c|}{\ding{55}}                                                      & 30k                                                                                \\ \cline{2-10} 
\multicolumn{1}{c|}{}                                                                               & ACKTR                 & \multicolumn{1}{c|}{14}                                                         & 2k                                                                                  & \multicolumn{1}{c|}{\ding{55}}                                                      & 30k                                                                                 & \multicolumn{1}{c|}{75}                                                         & 30k                                                                                 & \multicolumn{1}{c|}{\ding{55}}                                                      & 30k                                                                                \\ \hline
\multicolumn{1}{c|}{\multirow{3}{*}{\begin{tabular}[c]{@{}c@{}}LSTM-based\\ policy\end{tabular}}}   & ACER                  & \multicolumn{1}{c|}{12}                                                         & 1k                                                                                  & \multicolumn{1}{c|}{\ding{55}}                                                      & 30k                                                                                 & \multicolumn{1}{c|}{75}                                                         & 30k                                                                                 & \multicolumn{1}{c|}{\ding{55}}                                                      & 30k                                                                                \\ \cline{2-10} 
\multicolumn{1}{c|}{}                                                                               & A2C                   & \multicolumn{1}{c|}{12}                                                         & 1k                                                                                  & \multicolumn{1}{c|}{\ding{55}}                                                      & 30k                                                                                 & \multicolumn{1}{c|}{75}                                                         & 30k                                                                                 & \multicolumn{1}{c|}{\ding{55}}                                                      & 30k                                                                                \\ \cline{2-10} 
\multicolumn{1}{c|}{}                                                                               & ACKTR                 & \multicolumn{1}{c|}{12}                                                         & 1k                                                                                  & \multicolumn{1}{c|}{\ding{55}}                                                      & 30k                                                                                 & \multicolumn{1}{c|}{75}                                                         & 30k                                                                                 & \multicolumn{1}{c|}{\ding{55}}                                                      & 30k                                                                               
\end{tabular}%
}

\end{table}

%% file: sections/relatedWork.tex
In recent years, different forms of automata learning have been applied in combination with RL. Automata learning can aid RL by providing a stateful memory. This memory can be exploited either to further differentiate environment states or to capture steps required for
non-Markovian rewards.



Early work closely related to our approach is~\cite{DBLP:conf/aaai/Chrisman92,DBLP:conf/icml/McCallum93}. Both techniques
combine model learning and Q-learning for RL under partial observability, 
but they place stricter assumptions
on the environment, like knowledge about the number of environmental states.
More recently, Toro Icarte et al.~\cite{DBLP:conf/nips/IcarteWKVCM19} 
described optimization-based learning of finite-state models, called reward machines, to aid RL.
However, they require a labeling function on observations that meets certain
criteria and generally cannot handle changes in the transition 
probabilities, when observations stay the same.
DeepSynth~\cite{DBLP:conf/aaai/HasanbeigJAMK21} follows a similar approach, 
but focuses on sparse rewards
rather than partial observability. They learn automata via satisfiability 
checking to provide structure to complex tasks, where they also impose
requirements on a labeling function. 

Learning of reward machines has also been proposed to enable 
RL with non-Markovian rewards~\cite{DBLP:conf/aips/0005GAMNT020,DBLP:conf/cdmake/XuWONT21,DBLP:conf/aaai/NeiderGGT0021}, where
the gained rewards depend on the history of experiences rather than 
the current state and actions. In this context,
different approaches to automata learning are
applied to learn Mealy machines that keep track of previous
experiences in an episode. 
Velasquez et al.~\cite{DBLP:journals/corr/abs-2107-04633} extend reward-machine learning to a setting with stochastic non-Markovian rewards.
Our approach could be extended to non-Markovian rewards by
adding rewards to the observations.  
Subgoal automata inferred by Furelos-Blanco et al.~\cite{DBLP:conf/aaai/Furelos-BlancoL20} through answer set programming
serve a similar purpose as reward machines, by capturing interaction sequences
that need to occur for the successful completion of a task.  
Brafman et al.~\cite{DBLP:conf/aaai/GaonB20} learn deterministic finite
automata that also
encode which interactions lead to a reward in RL with non-Markovian
rewards. Similarly to our approach, the states of learned automata are used
as additional observations.

%% file: sections/conclusion.tex
We propose an approach for reinforcement learning under partial observability.
For this purpose, we combine Q-learning with the automata learning 
technique \IOALERGIA. With automata learning, we learn hidden information
about the environment's state structure that provides additional observations
to Q-learning, thus enabling this form of learning in POMDPs. 
We evaluate our approach in partially observable environments and show that
it can outperform the baseline deep RL approach with LSTMs and fixed memory. 

For future work, we plan to generalize our approach to other (deep) 
RL approaches by integrating explored learned MDP states as observations. 
Approaches already including experience replay naturally lend themselves to 
such extensions, since we merely need to change the replay mechanism 
and execute it after updating the learned MDP. 
To scale the proposed approach to larger environments, 
we intend to explore and develop automata-learning techniques that can model high-dimensional data.